%% file: main.tex
\documentclass[10pt,twocolumn,letterpaper]{article}

 \usepackage[pagenumbers]{cvpr} 


\definecolor{cvprblue}{rgb}{0.21,0.49,0.74}
\usepackage[pagebackref,breaklinks,colorlinks,allcolors=cvprblue]{hyperref}

\usepackage[utf8]{inputenc} 
\usepackage[T1]{fontenc}    
\usepackage{url}            
\usepackage{booktabs}       
\usepackage{amsfonts}       
\usepackage{amssymb}     
\usepackage{amsmath}     
\usepackage{nicefrac}       
\usepackage{microtype}      
\usepackage{xcolor}         
\usepackage{bm}     
\usepackage{xfrac} 
\usepackage{array}
\usepackage{makecell}
\usepackage{multirow}
\usepackage{wrapfig}
\usepackage{enumitem}
\usepackage{caption}
\usepackage{algorithm,algorithmicx,algpseudocode}

\newcommand{\PreserveBackslash}[1]{\let\temp=\\#1\let\\=\temp}
\newcolumntype{C}[1]{>{\PreserveBackslash\centering}p{#1}}
\newcolumntype{R}[1]{>{\PreserveBackslash\raggedleft}p{#1}}
\newcolumntype{L}[1]{>{\PreserveBackslash\raggedright}p{#1}}

\def\paperID{6314} 
\def\confName{CVPR}
\def\confYear{2025}

\title{PatchScaler: An Efficient Patch-Independent Diffusion \\ Model for Image Super-Resolution}

\author{%
	Yong Liu$^{1,2}$ \qquad
	Hang Dong$^{3}$\thanks{Corresponding author.} \qquad
	Jinshan Pan$^{4}$ \qquad
	Qingji Dong$^{1,2}$ \qquad
	Kai Chen$^{3}$ \\
	Rongxiang Zhang$^{3}$ \qquad
	Lean Fu$^{3}$ \qquad
	Fei Wang$^{1,2}$\\
	{$^{1}$National Key Laboratory of Human-Machine Hybrid Augmented Intelligence} \\
	{$^{2}$IAIR, Xi’an Jiaotong University} \qquad
	{$^{3}$ByteDance Inc} \\
	{$^{4}$Nanjing University of Science and Technology}
	\vspace{-0.2cm}}

\begin{document}
\maketitle
\begin{abstract}
While diffusion models significantly improve the perceptual quality of super-resolved images, they usually require a large number of sampling steps, resulting in high computational costs and long inference times. 
Recent efforts have explored reasonable acceleration schemes by reducing the number of sampling steps. 
However, these approaches treat all regions of the image equally, overlooking the fact that regions with varying levels of reconstruction difficulty require different sampling steps. 
To address this limitation, we propose PatchScaler, an efficient patch-independent diffusion pipeline for single image super-resolution. 
Specifically, PatchScaler introduces a Patch-adaptive Group Sampling (PGS) strategy that groups feature patches by quantifying their reconstruction difficulty and establishes shortcut paths with different sampling configurations for each group. 
To further optimize the patch-level reconstruction process of PGS, we propose a texture prompt that provides rich texture conditional information to the diffusion model. 
The texture prompt adaptively retrieves texture priors for the target patch from a common reference texture memory. 
Extensive experiments show that our PatchScaler achieves superior performance in both quantitative and qualitative evaluations, while significantly speeding up inference. 
Our code will be available at~\url{https://github.com/yongliuy/PatchScaler}. 
\end{abstract}


\section{Introduction}
Single image super-resolution (SISR) aims to reconstruct a high-resolution (HR) image from its low-resolution (LR) observation, 
a highly ill-posed problem due to the unknown degradation processes in real-world scenarios. 
As two predominant approaches, deep convolutional neural network (CNN)-based~\cite{zhou2020cross, li2020lapar, wang2021real, zhang2024real} and Transformer-based~\cite{zhang2022efficient, liu2023unfolding, liang2021swinir, yang2021implicit} SISR methods have made significant progress in the past decade. 
However, most of them are optimized primarily for the peak signal-to-noise ratio (PSNR) and the structural similarity index measure (SSIM), which are less effective in restoring realistic image details. 

Diffusion models \cite{sohl2015deep, ho2020denoising}, with their powerful ability to model data distributions from noise, have offered immense potential for conditional generation tasks, such as image synthesis \cite{dhariwal2021diffusion, rombach2022high, yang2024improving}, image editing \cite{brooks2023instructpix2pix, kawar2023imagic, yang2024dynamic}, and image super-resolution \cite{wang2024exploiting, yang2023pixel, lin2023diffbir}. 
Although diffusion models have achieved significant success in the field of image super-resolution (SR), their inference process remains inefficient due to the large number of iterative sampling steps required for high-quality image reconstruction. 
\begin{figure}[t]
	\centering
	\includegraphics[width=0.92\linewidth]{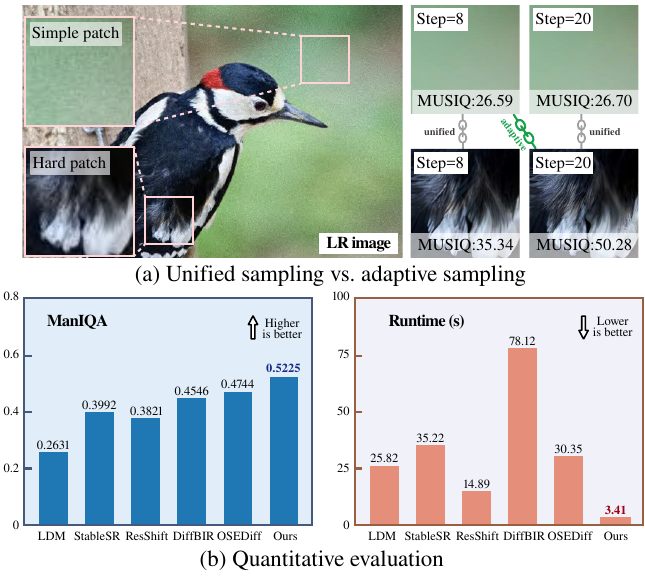}
	\vspace{-2mm}
	\caption{(a) Qualitative analysis of unified sampling and adaptive sampling.  (b) Quantitative comparison of diffusion-based SR methods on RealSR \cite{cai2019toward} dataset. Noted that the runtime is measured on the $ \times $4 (512 $ \to  $2048) SR task using an NVIDIA Tesla A100 GPU.}
	\label{fig_observation}
	\vspace{-6mm}
\end{figure}

Recent methods have attempted to accelerate diffusion-based SR models by introducing conditional distillation \cite{mei2023conditional, wang2023sinsr} or redefining the diffusion process \cite{yue2024resshift, yue2024efficient}. 
However, these approaches tend to treat all regions of an image equally and indiscriminately reduce the number of sampling steps, which inevitably compromises the quality of the super-resolved images. 
Furthermore, when dealing with high-resolution images, these approaches continue to incur substantial computational costs. 
In this paper, we observe that a unified sampling process is suboptimal for all patches of an image. 
As illustrated in \Cref{fig_observation} (a), patches with fewer structural details can be effectively reconstructed with fewer sampling steps, whereas patches rich in textural information require more sampling steps. 
This observation motivates us to develop a patch-adaptive accelerated diffusion model for SISR. 
Based on this insight, we propose PatchScaler, an efficient diffusion-based SR method that dynamically accelerates the inference process through an adaptive, patch-independent diffusion pipeline.  
Specifically, we first employ a global restoration module to generate a coarse HR feature along with a confidence map that reflects the reconstruction difficulty across different regions. 
We then introduce Patch-adaptive Group Sampling (PGS), which divides the coarse HR features into patches and groups them according to quantified reconstruction difficulty. 
PGS identifies an optimal intermediate point and sampling configurations for each group, enabling a shortcut path from coarse HR patches to the ground truths. 
Finally, we introduce a Patch-wise Diffusion Transformer~(Patch-DiT) as the backbone of PatchScaler to refine the fine textures from the coarse HR patches. 
Since the reconstruction of local image textures plays a more critical role for SISR than the Text2Image task, Patch-DiT is naturally well-suited for restoring lost details. 
To further optimize the patch-level reconstruction process of PGS, we propose a texture prompt that supplies rich conditional information to Patch-DiT by retrieving high-quality texture priors for the target patch from a universal reference texture memory. 
The texture prompt is effective for reconstructing local details and addressing the misalignment issue between image and text prompts in diffusion models~\cite{chen2023pixart}. 
Experiments show that our PatchScaler significantly accelerates inference speed (e.g., only 0.23$\times$ the runtime of ResShift \cite{yue2024resshift} on the 512$ \to  $2048 SR task), while still maintaining superior performance, as shown in \Cref{fig_observation}(b). 
We summarize our main contributions as follows: 
\begin{itemize}
	\item We propose PatchScaler, a novel patch-independent SR pipeline that employs patch-adaptive group sampling to dynamically accelerate the sampling process, enabling efficient restoration of high-resolution images. 
	\item We present an effective texture prompt for the Patch-wise Diffusion Transformer (Patch-DiT) in PatchScaler to provide rich texture priors and improve reconstruction quality. 
	\item Experiments show that PatchScaler achieves favorable performance on several datasets against state-of-the-art SR methods and is much more efficient than previous diffusion-based SR methods. 
\end{itemize}
%
%
\section{Related Work}
\textbf{Single Image Super-Resolution.} 
Over the past decade, a series of deep networks~\cite{li2018multi, ning2021uncertainty, sun2022shufflemixer, zhou2020cross, liu2023unfolding} have been proposed to address SISR challenges. 
For real-world degradation scenarios, 
Ji \textit{et al.}~\cite{ji2020real} proposed RealSR that learns the specific degradation of blurry and noisy images by estimating the kernel and noise. 
By incorporating GANs \cite{goodfellow2020generative, schonfeld2020u, wang2018esrgan} into the SISR task, Zhang~\textit{et al.} \cite{zhang2021designing} presented BSRGAN, a practical degradation model that synthesizes realistic degradations via a random shuffling strategy. 
Wang \textit{et al.}~\cite{wang2021real} proposed Real-ESRGAN that employs a high-order degradation process to simulate practical degradations and incorporates sinc filters to replicate common ringing and overshoot artifacts. 
Liang \textit{et al.}~\cite{liang2021swinir} introduced SwinIR-GAN based on the Swin Transformer~\cite{liu2021swin}, to achieve competitive image restoration performance. 
Nevertheless, these approaches often struggle to generate realistic fine details. 
\noindent\textbf{Diffusion Model.} 
The powerful generative capabilities inherent in diffusion models have yielded remarkable performance in SISR task. 
Specifically, Wang \textit{et al.} \cite{wang2024exploiting} presented StableSR, integrating LR images into diffusion models via a time-aware encoder to reconstruct high-quality HR images.
Lin \textit{et al.} \cite{lin2023diffbir} proposed a unified restoration framework DiffBIR, which sequentially uses two stages of restoration and generation to ensure fidelity and realism. 
Yue \textit{et al.} \cite{yue2024resshift} introduced ResShift, utilizing a novel iterative sampling approach from LR to HR images by shifting residuals. 
Although these approaches have achieved considerable improvements in perceptual quality, they treat all regions of an image equally and indiscriminately adopt a unified sampling strategy during inference, even for easily restored patches. 
In contrast, our work explores a more efficient patch-independent diffusion pipeline, achieving superior results with faster inference speed. 
%

	
%
\section{Proposed Method}

\begin{figure*}[t]
	\centering
	\includegraphics[width=0.98\linewidth]{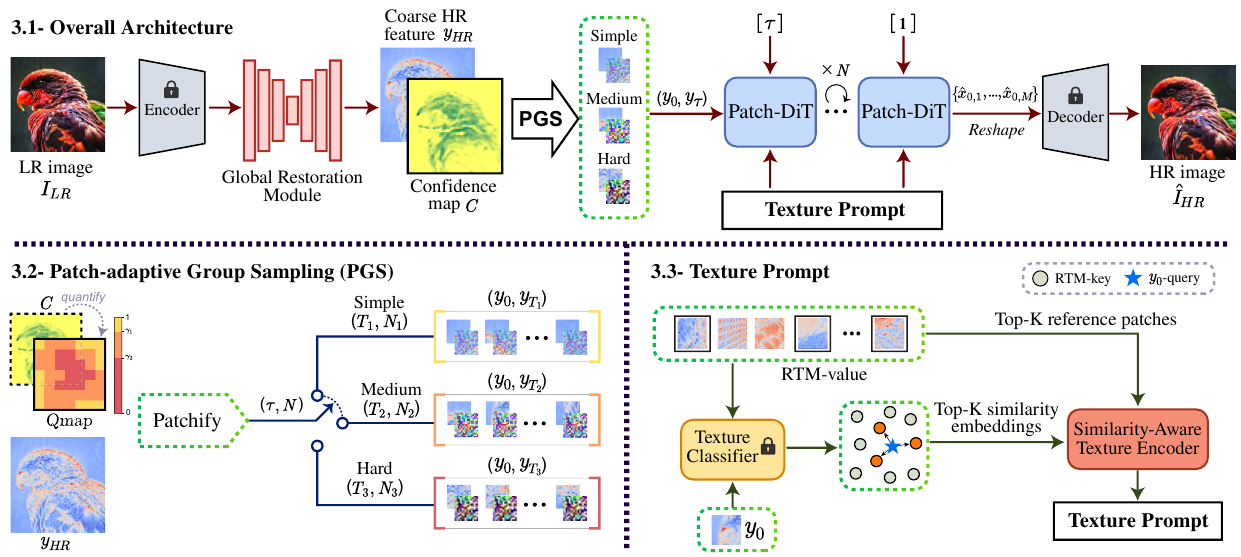}
	\caption{Overview of the proposed PatchScaler. PGS dynamically assign feature patches into groups with different sampling configuration based on quantified confidence map. Moreover, the texture prompt provides high-quality conditional information for Patch-DiT by retrieving high-quality texture priors from universal RTM.} 
	\label{fig_network} 
	\vspace{-4mm}
\end{figure*}
\subsection{Overall Architecture}
\label{sec_overall_architecture}
The overall architecture of PatchScaler is illustrated in \Cref{fig_network}. 
Given an LR image $ I_{LR} \in \mathbb{R}^{3\times H\times W} $, we first encode it into a latent representation $ \boldsymbol{y}_\mathit{LR} \in \mathbb{R}^{c\times \frac{H}{d}\times \frac{W}{d}} $ through the frozen encoder of the autoencoder,
where $ d $ is the downsampling factor. 
Next, we employ a Global Restoration Module (GRM) to remove the degradations (e.g., noise or distortion artifacts) and capture long-range dependencies in $ \boldsymbol{y}_\mathit{LR} $. 
To assess the reconstruction difficulty of different regions, a coarse HR feature $ \boldsymbol{y}_\mathit{HR} \in \mathbb{R}^{c\times \frac{H}{d}\times \frac{W}{d}} $ and a corresponding confidence map $ C \in \mathbb{R}^{1\times \frac{H}{d}\times \frac{W}{d}} $ are simultaneously generated by GRM. 
Thus, we incorporate the confidence-driven loss \cite{ning2021uncertainty} as a constraint and the training objective is: 
\begin{equation}
 L(\theta):=\left\| \boldsymbol{y}_\mathit{HR} - \boldsymbol{x}_\mathit{HR} \right\|_{1}^{2} \\ 
 +\lambda(C\left\| \boldsymbol{y}_\mathit{HR} - \boldsymbol{x}_\mathit{HR} \right\|_{2}^{2} - \eta \log(C)). 
 \label{grm_loss}
\end{equation} 

\begin{figure}[t]
	\centering
	\includegraphics[width=0.45\textwidth]{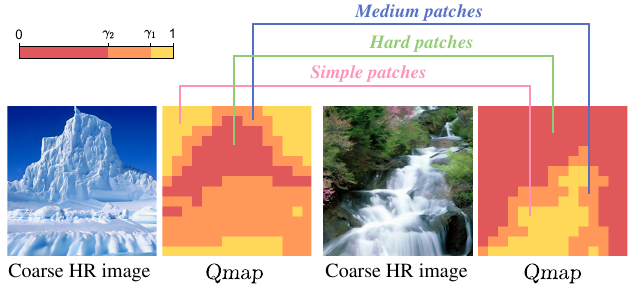}
    \vspace{-2mm}
	\caption{Examples of coarse HR images and corresponding $Qmap$. Our approach can accurately quantify the reconstruction difficulty of different regions across diverse scenes.} 
	\label{fig_quantify} 
	\vspace{-4mm}
\end{figure}

Before initiating the diffusion process, we introduce Patch-adaptive Group Sampling (PGS), which dynamically divides $ \boldsymbol{y}_\mathit{HR} $ into patches grouped according to their reconstruction difficulties. 
Specifically, PGS partitions $ \boldsymbol{y}_\mathit{HR} $ into a patch set $ \{\boldsymbol{y}_{0,1}, \boldsymbol{y}_{0,2}, ..., \boldsymbol{y}_{0,M}\} $, where $ \boldsymbol{y}_{0,i} \in \mathbb{R}^{c\times V \times V} $, with $ V $ and $ M $ denote the size of each patch and the length of the patch set, respectively.
To define the reconstruction difficulty of $ \boldsymbol{y}_{0,i} $, a quantified confidence map $ Qmap $ is generated by averaging the confidence map $ C $ within the patch:
\begin{equation}
    Qmap_{\boldsymbol{y}_{0,i}}:=\left\{\begin{array}{ll}
    \text {Simple}, & Avg(C\left \langle \boldsymbol{y}_{0,i}\right \rangle ) \in (\gamma_1 ,1], \\
    \text {Medium}, & Avg(C\left \langle \boldsymbol{y}_{0,i}\right \rangle ) \in (\gamma_2 ,\gamma_1], \\
    \text {Hard}, & Avg(C\left \langle \boldsymbol{y}_{0,i}\right \rangle ) \in [0 ,\gamma_2]
    \end{array}\right.
    \label{gamma}
\end{equation} 
where $ Avg(C\left \langle \boldsymbol{y}_{0,i}\right \rangle) $ represents the average confidence value over the patch, $ \gamma_1 $ and $ \gamma_2 $ are the confidence thresholds. 
Based on the resulting $ Qmap $, PGS classifies the patches into different groups (i.e., "simple", "medium", and "hard") and assigns optimized sampling configurations during the subsequent diffusion process. 
\Cref{fig_quantify} presents two examples of coarse HR images generated by GRM along with the corresponding quantified confidence maps $Qmap$ calculated by Equation~\eqref{gamma}. 
It can be seen that our method accurately quantifies the reconstruction difficulty of different regions within the image. 
Patches with complex textures are assigned to the "hard" group, while relatively smooth patches are categorized as the "simple" group. 
In this way, PGS provides additional flexibility in handling various feature regions, thus significantly accelerating the inference process. 
Since most popular open-source diffusion models~\cite{rombach2022high, meng2023distillation, ruiz2023dreambooth} produce inferior results in low-resolution patches~\cite{zheng2024any}, we build the diffusion backbone, Patch-DiT, based on DiTs\cite{peebles2023scalable}. 
Due to the inherent token sequence designation of Transformer, Patch-DiT is well-suited to handle patch-level features. 
Furthermore, we introduce a novel texture prompt as a conditional input for Patch-DiT to enhance its texture generation capabilities. 
The texture prompt provides rich texture priors by retrieving high-quality texture patches from a universal reference texture memory. 
Finally, the HR patch set $ \{\hat{\boldsymbol{x}}_{0,1}, \hat{\boldsymbol{x}}_{0,2}, ..., \hat{\boldsymbol{x}}_{0,M}\} $ generated by Patch-DiT is reshaped into $ \hat{\boldsymbol{x}}_{HR} \in \mathbb{R}^{4\times \frac{H}{d}\times \frac{W}{d}} $ and passed to the decoder to obtain the HR output $ \hat{\boldsymbol{I}}_\mathit{HR} $.
%
%

%
\subsection{Patch-adaptive Group Sampling}
\begin{figure}[t]
    \centering
    \resizebox{0.26\textwidth}{!}{
        \includegraphics[width=0.26\textwidth]{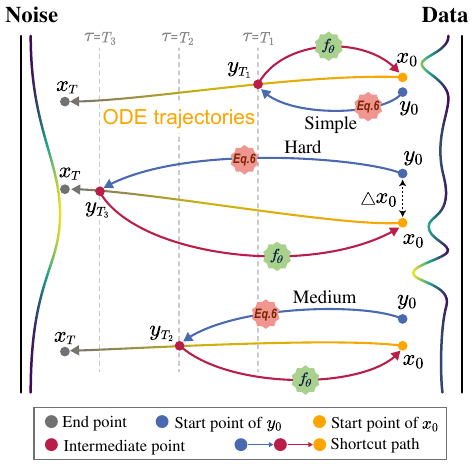}
    }
    \hfill 
    \raisebox{52pt}{
    \begin{minipage}[c]{0.20\textwidth}
        \caption{Illustration of the proposed PGS, which establishes a shortcut path between $ \boldsymbol{x}_{0} $ and $ \boldsymbol{y}_{0} $. PGS treats different patches discriminatively and dynamically assign different sampling configurations. Here, $ \boldsymbol{y}_{0} $ and $ \boldsymbol{x}_{0} $ denote the coarse HR patch and ground truth, respectively. $ f_{\theta } $ denotes the diffusion model Patch-DiT.}
	   \label{fig_pas}
    \end{minipage}
    }
    \vspace{-8mm}
\end{figure}

Existing text-to-image models, such as Denoising Diffusion Probabilistic Models (DDPM) \cite{ho2020denoising, sohl2015deep}, assume a forward diffusion trajectory where Gaussian noise is gradually applied to real data $ \boldsymbol{x}_{0} $:
\begin{equation}
	q\left(\boldsymbol{x}_{t} \mid \boldsymbol{x}_{t-1}\right)=\mathcal{N}\left(\boldsymbol{x}_{t} ; \sqrt{1-\beta_{t}} \boldsymbol{x}_{t-1}, \beta_{t} \mathbf{I}\right), 
\end{equation} 
where $ t=1, \ldots, T $ and $ \boldsymbol{x}_{t} $ is the variable at time step $ t $. 
The noise schedule $ \left\{\beta_{t}\right\} $ can either be predefined or learned. 
Using the reparameterization trick, this process can be expressed in a closed form as: 
\begin{equation}
	q\left(\boldsymbol{x}_{t} \mid \boldsymbol{x}_{0}\right) =\mathcal{N}\left(\boldsymbol{x}_{t} ; \sqrt{\bar{\alpha}_{t}} \boldsymbol{x}_{0},\left(1-\bar{\alpha}_{t}\right) \mathbf{I}\right), 
	\label{eq_diffusion}
\end{equation} 
where $ \alpha_{t}:=1-\beta_{t} $ and $ \bar{\alpha}_{t}:=\prod_{i=0}^{t} \alpha_{i} $. 
$ \boldsymbol{x}_{t} $ can be sampled by $ \boldsymbol{x}_{t}=\sqrt{\bar{\alpha}_{t}} \boldsymbol{x}_{0}+\sqrt{\left(1-\bar{\alpha}_{t}\right)} \boldsymbol{\epsilon} $ for $ \boldsymbol{\epsilon} \sim \mathcal{N}(0, \boldsymbol{I}) $. 
In the reverse process, denoising models $ \boldsymbol{f}_{\theta}\left(\boldsymbol{x}_{t}, t\right) $ are trained to map any point at any time step on the diffusion trajectory to the starting point $ \boldsymbol{x}_{0} $. 
However, when these diffusion models are applied to SISR, there is a deviation between the LR image and the ground truth. 
To obtain a reasonable initial point for the reverse denoising process, 
previous approaches either apply a large noise strength ($i.e., t=T$) to the LR image, approximating the endpoint $ \boldsymbol{x}_{T} $ of diffusion trajectory, or sample $ \boldsymbol{x}_{T} $ directly from a Gaussian noise $ \mathcal{N}(0, \mathbf{I}) $. 
Subsequently, the HR result is reconstructed step by step from $ \boldsymbol{x}_{T} $ with a unified global sampling that treats all regions of the image equally. 
However, we argue that it is redundant because simple patches can be reconstructed with fewer iterations than hard patches, as illustrated in \Cref{fig_observation}(a).
To enable dynamic acceleration of the denoising process, we propose Patch-adaptive Group Sampling (PGS), which establishes a shortcut path between HR patch $ \boldsymbol{x}_{0} $ and coarse HR patch $ \boldsymbol{y}_{0} $, as shown in \Cref{fig_pas}. 
Here, $ \boldsymbol{y}_{0} $ is taken from the patch set. 

Specifically, let $ \boldsymbol{x}_{0}= \boldsymbol{y}_{0}+\triangle\boldsymbol{x}_{0} $, then Equation \eqref{eq_diffusion} can be reformulated as: 
\begin{equation}
	q\left(\boldsymbol{x}_{t} \mid \boldsymbol{y}_{0}, \triangle\boldsymbol{x}_{0}\right) =\mathcal{N}\left(\boldsymbol{x}_{t} ; \sqrt{\bar{\alpha}_{t}} (\boldsymbol{y}_{0}+\triangle\boldsymbol{x}_{0}),\left(1-\bar{\alpha}_{t}\right) \mathbf{I}\right). 
	\label{pas_diffusion2}
\end{equation} 
The goal of PGS is to identify an appropriate intermediate point along the diffusion trajectory for each patch based on the distance $ \triangle\boldsymbol{x}_{0} $. 
Since the GRM has removed the degradation (such as noise and distortion artifacts) from the LR input, $ \triangle\boldsymbol{x}_{0} $ is relatively small, especially for patches with fewer structural details. 
Therefore, an appropriate intermediate time step $ \tau \in [1, T] $ can be determined such that $ \sqrt{\bar{\alpha}_{\tau}}\triangle \boldsymbol{x}_{0} \to 0 $. 
For small $ \triangle\boldsymbol{x}_{0} $, $ \tau $ can be small enough, while for a larger values $ \triangle\boldsymbol{x}_{0} $, a larger $ \tau $ is required to ensure that $ \bar{\alpha}_{\tau} $ is sufficiently small. 
In this way, we can approximate the original diffusion trajectory at time step $ \tau $ as: 
\begin{equation}
	q\left(\boldsymbol{x}_{\tau} \mid \boldsymbol{y}_{0}\right) \approx \mathcal{N}\left(\boldsymbol{x}_{\tau} ; \sqrt{\bar{\alpha}_{\tau}} \boldsymbol{y}_{0},\left(1-\bar{\alpha}_{\tau}\right) \mathbf{I}\right).  
	\label{pas_diffusion3}
\end{equation} 
This finding encourages us to introduce a new forward diffusion trajectory for the coarse HR patch $ \boldsymbol{y}_{0} $ with truncated forward diffusion time steps: 
\begin{equation}
	q\left(\boldsymbol{y}_{t} \mid \boldsymbol{y}_{0}\right) =\mathcal{N}\left(\boldsymbol{y}_{t} ; \sqrt{\bar{\alpha}_{t}} \boldsymbol{y}_{0},\left(1-\bar{\alpha}_{t}\right) \mathbf{I}\right), 
	\label{pas_diffusion}
\end{equation}  
where $ t=1, \ldots, \tau $. 
As a result, the reverse process can start from a non-Gaussian distribution $ \boldsymbol{y}_{\tau} $ that has a shorter distance to $ \boldsymbol{x}_{0} $, thus significantly reducing the required number of sampling steps $ N $. 
This allows PGS to discriminatively handle feature patches with varying reconstruction difficulties, thereby increasing the flexibility of the reverse denoising process. 
In practice, $ \triangle\boldsymbol{x}_{0} $ is estimated based on the confidence map $C$ as the ground truth $ \boldsymbol{x}_{0} $ is unavailable during inference. 
To facilitate parallel computing and improve efficiency, we derive the quantified confidence map $ Qmap $ from $ C $ using Equation \eqref{gamma} and divide the coarse HR feature patches into three groups (i.e., simple, medium, and hard), as shown in the bottom-left part of \Cref{fig_network}. 
Furthermore, we set different intermediate time steps $ T_1 < T_2 < T_3 $ and sampling steps $ N_1<N_2<N_3 $ for patches belonging to "simple", "medium", and "hard" groups, respectively.
By assigning different sampling configurations to each group, PatchScaler can dynamically accelerate the sampling process. 

\begin{algorithm}[tb]
    \caption{Retrieval Texture Priors from RTM.} 
        \label{alg:tp}
    \begin{algorithmic}[1]
    
        \Statex \hspace{-1.5em}\textbf{Input:} $\mathcal{T}$; RTM-\textit{key}; RTM-\textit{value}; $ \boldsymbol{y}_{0} $ 
    
        \hspace{-3.5em}\Comment{$\mathcal{T}$ represents the feature extraction module of Texture Classifier; RTM-\textit{key}=Normalize($\mathcal{T}$(RTM-\textit{value}))}.
        
        \State Obtain corresponding semantic feature vector from $ \boldsymbol{y}_{0} $: 
    
        \hspace{-2.0em} $ \boldsymbol{y}\textit{-query}=\mathcal{T}(\boldsymbol{y}_{0} ) $; 
    
        \State Calculate normalized inner products between $ \boldsymbol{y}_{0}\text{-\textit{query}} $ and $ \text{RTM-\textit{key}}_i $: 
    
        \hspace{-2.0em} $s_i=\left \langle \text{RTM-\textit{key}}_i,\text{Normalize}(\boldsymbol{y}_{0}\text{-\textit{query}})  \right \rangle$; 
    
        \State Ranking $ s_i $ and determine top-K indexs $ d^k $ and top-K similarities $ s^k $;
    
        \State Retrieve corresponding texture prior $ tp^k $ from RTM-\textit{value} based on top-K indexs $ d^k $;
    
        \Statex \hspace{-1.5em}\textbf{Output:} Texture prior $ tp^k $; top-K similarities $ s^k $. 
    
    \end{algorithmic}
\end{algorithm}

\subsection{Texture Prompt}
\label{sec:texture_prompt}
To optimize the patch-level reconstruction process of PGS, we propose a texture prompt inspired by the reference SR \cite{zhang2019image, yan2020towards, qin2023blind}, which provides rich conditional information for Patch-DiT. 
The texture prompt retrieves highly correlated texture priors from a Reference Texture Memory (RTM) and transfers them into the mainstream of Patch-DiT, as illustrated in the bottom-right part of \Cref{fig_network}. 
\noindent\textbf{Reference Texture Memory.} 
Our RTM serves as a repository for storing diverse and high-quality texture feature patches (denoted as RTM-\textit{value}), along with their corresponding semantic feature vectors (denoted as RTM-\textit{key}), to provide universal texture priors for Patch-DiT. 
To construct the RTM, we first extract 20,000 initial latent texture patches (i.e., RTM-\textit{value}) from the high-quality image datasets DIV2K \cite{agustsson2017ntire}, OutdoorSceneTraining \cite{wang2018recovering}, and Manga109 \cite{matsui2017sketch}, which contain a wide variety of categories, such as animal, sky, cartoon, building, mountain, and plant. 
Subsequently, we introduce a pre-trained Texture Classifier to extract deep semantic representation (i.e., RTM-\textit{key}) from the initial RTM-\textit{value}. 
To accelerate the feature retrieval process, we employ the farthest point sampling algorithm \cite{qi2017pointnet} on both RTM-\textit{value} and RTM-\textit{key} to sample the final 2,000 representative pairs, which form our final RTM. 
It is important to note that all of these operations are performed offline, thus incurring no additional inference cost. 
\noindent\textbf{Texture Retrieval and Transfer.} 
Our objective is to retrieve high-quality texture priors with semantic similarity to the target feature patch $ \boldsymbol{y}_{0} $ from RTM and transfer them into Patch-DiT as a texture prompt. 
To achieve this, we first use the texture classifier to project $ \boldsymbol{y}_{0} $ into a semantic feature vector, denoted as $ \boldsymbol{y}_{0}$-$\textit{query} $. 
We then calculate the similarity scores between $ \boldsymbol{y}_{0}$-$\textit{query} $ and RTM-\textit{key} using normalized inner products \cite{zhang2019image}. 
By ranking these similarity scores, we retrieve the top-\textit{K} matches, denoted as $ d^k $, along with corresponding texture prior $ tp^k $, for $ \boldsymbol{y}_{0}$-$\textit{query} $ across the entire RTM-\textit{key}. 
The complete retrieval process is shown in Algorithm~\ref{alg:tp}. 
Subsequently, we integrate a similarity-aware texture encoder to transfer texture priors into Patch-DiT as a texture prompt. 
The texture prompt is introduced into each DiT block of Patch-DiT via a cross-attention mechanism \cite{vaswani2017attention}, providing rich texture conditional guidance for the denoising process. 
Notably, our similarity-aware texture encoder is time-step independent, meaning that it does not need to be recalculated for each iteration, significantly accelerating inference process.  
As an efficient alternative, we introduce dimension-wise scaling parameters based on the time embedding in each cross-attention layer. 
Furthermore, by incorporating texture prompt, we mitigate the text-image misalignment issues typical in diffusion models with text prompt \cite{chen2023pixart}, which is particularly beneficial for SISR, as it ensures more consistent results in image super-resolution. 
%

%
\section{Experiments}

\subsection{Experimental Settings}
\textbf{Training Details}. 
We train our PatchScaler based on the LSDIR~\cite{li2023lsdir} and HQ-50K \cite{yang2023hq} datasets, which consist of large-scale high-resolution images for image restoration. 
During training, the corresponding LR images are synthesized using the degradation pipeline of Real-ESRGAN~\cite{wang2021real}. 
The patch size $V$ is set to 16. 
We utilize the Adam optimizer \cite{kingma2014adam} with a batch size of 6, and the learning rate is fixed as 5e-5. 
Our PatchScaler is trained for 700K iterations using the Adam optimizer \cite{kingma2014adam} with a batch size of 6, while GRM only participates in the training for the first 100K iterations based on Equation~\eqref{grm_loss}. 
The training process takes 10 days using eight NVIDIA Tesla V100 GPUs. 
More details about the network architecture are included in the supplementary material. 
\noindent\textbf{Inference Details.}
We evaluate our PatchScaler on both synthetic and real-world datasets.
For synthetic data, we generate LR-HR pairs based on randomly selected 2000 test images from LSDIR~\cite{li2023lsdir}, referred to as LSDIR-Test, following the degradation method used in ResShift \cite{yue2024resshift}. 
We also evaluate our model on two real-world datasets: RealSR~\cite{cai2019toward} and RealSet110. 
As the degradation in RealSR is monotonic and lacks common SR scenes such as animals and illustrations, we collect an extended dataset named RealSet110. 
It is worth noting that 98 of the LR images were collected from previous works~\cite{lugmayr2020ntire, zhang2021designing, yue2024resshift, karras2019style, bell2015material, yang2023hq, lin2023diffbir}, with the remaining images were collected from online sources. 
To evaluate the performance of our PatchScaler at both patch and image scales, the resolution of the LR images in LSDIR-Test and RealSR datasets is fixed at 128$\times$128, while the images in RealSet110 dataset have arbitrary resolutions. 
We classify the patches into three groups—"simple", "medium", and "hard"—following previous work \cite{kong2021classsr, wang2024camixersr, wang2023classification}, and argue that increasing the number of groups would make the differences between patches with different levels of reconstruction difficulty less obvious and increase the inference complexity of the model. 
Furthermore, to eliminate potential boundary effects within patches, we decompose the coarse HR feature into overlapping patches, which is commonly used in leading diffusion-based methods \cite{bar2023multidiffusion, wang2024exploiting}. 
We also apply wavelet-based color normalization \cite{wang2024exploiting} on the super-resolved image to align its low-frequency features with those of the LR input. 
\begin{table*}[t]
	\centering
	\caption{Quantitative comparison with state-of-the-art SR methods on both synthetic and real-world datasets. The best and second best results are \textcolor{red!60!black}{\textbf{highlighted}} and \underline{underlined}, respectively. PSNR/SSIM on Y channel are reported on each dataset. We calculate ManIQA metric based on the official ManIQA-KONIQ pre-trained model. 
    }
	\label{tab_quantitative_results}
	\small
	\vspace{-2mm}
        \resizebox{0.98\textwidth}{!}{
	\begin{tabular}{l|cccccc|ccc|ccc}
		\Xhline{1.2pt}
		\multirow{2}{*}{Methods} & \multicolumn{6}{c}{LSDIR-TEST} & \multicolumn{3}{|c}{RealSR} & \multicolumn{3}{|c}{RealSet110}  \\ \cline{2-13} 
		& PSNR & SSIM & LPIPS & ManIQA & CLIPIQA & MUSIQ & ManIQA & CLIPIQA & MUSIQ & ManIQA & CLIPIQA & MUSIQ     \\ 
		\hline
		RealSR-JPEG~\cite{ji2020real}    & 22.09 & 0.4819  & 0.3982 & 0.3647 & 0.6466 & 62.47 & 0.1710 & 0.5267  & 33.69 & 0.3030 & 0.7588 & 54.03 \\
		BSRGAN~\cite{zhang2021designing} & 23.74 & \underline{0.5748} & 0.3336 & 0.4069 & 0.6853 & 69.09 & 0.3787 & 0.5512 & 63.20 & 0.3870 & 0.7918 & 67.13 \\
		Real-ESRGAN~\cite{wang2021real}    & 23.08 & \textcolor{red!60!black}{\textbf{0.5758}}  & 0.3234 & 0.4335 & 0.6810 & 69.91 & 0.3811 & 0.5492  & 60.56 & 0.3779 & 0.7844 & 65.05 \\
		SwinIR-GAN~\cite{liang2021swinir}    & 23.05 & 0.5698  & 0.3262 & 0.4195 & 0.6833 & 68.72 & 0.3583 & 0.5669  & 59.23 & 0.3598 & \underline{0.8009} & 63.92 \\
		LDM~\cite{rombach2022high}    & \textcolor{red!60!black}{\textbf{24.14}} & 0.5630 & 0.3323 & 0.3466 & 0.6735 & 61.83 & 0.2631 & 0.5617 & 47.72 & 0.2834 & 0.7628 & 55.01 \\
		StableSR~\cite{wang2024exploiting}    & 23.09 & 0.5664 & 0.3170 & 0.4772 & 0.6739 & 69.91 & 0.3992 & 0.5350 & 61.33 & 0.3856 & 0.7851 & 62.08 \\
		ResShift~\cite{yue2024resshift}    & \underline{23.75} & 0.5686 & \textcolor{red!60!black}{\textbf{0.3102}} & 0.5141 & 0.6919 & 69.62 & 0.3821 & 0.5694 & 58.81 & 0.4011 & 0.7584 & 62.07 \\
        PASD~\cite{yang2023pixel}    & 21.02 & 0.4940 & 0.3651 & 0.5177 & 0.6847 & 71.91 & 0.4656 & 0.5607 & 67.44 & 0.4378 & 0.7608 & 66.03 \\
		DiffBIR~\cite{lin2023diffbir}    & 23.33 & 0.5305 & 0.3469 & \underline{0.5406} & 0.6726 & 69.78 & 0.4546 & 0.5762 & 62.75 & \underline{0.4933} & 0.7756 & 66.75 \\
        SinSR~\cite{wang2023sinsr}    & 23.52 & 0.5577 & 0.3170 & 0.4866 & 0.6960 & 69.73 & 0.3998 & 0.5770 & 60.75 & 0.4250 & 0.6738 & 55.24 \\
        OSEDiff~\cite{wu2024one}    & 22.50 & 0.5357 & \underline{0.3103} & 0.4684 & \underline{0.7070} & \underline{72.16} & \underline{0.4744} & \textcolor{red!60!black}{\textbf{0.5890}} & \underline{69.12} & 0.4679 & 0.7949 & \underline{70.26} \\
        \hline
		PatchScaler   & 22.60 & 0.5297 & 0.3520 & \textcolor{red!60!black}{\textbf{0.6261}} & \textcolor{red!60!black}{\textbf{0.7145}} & \textcolor{red!60!black}{\textbf{73.23}} & \textcolor{red!60!black}{\textbf{0.5225}} & \underline{0.5839} & \textcolor{red!60!black}{\textbf{69.50}} & \textcolor{red!60!black}{\textbf{0.5442}} & \textcolor{red!60!black}{\textbf{0.8213}} & \textcolor{red!60!black}{\textbf{70.97}} \\
		\Xhline{1.2pt}
	\end{tabular}}
	\vspace{-2mm}
\end{table*}
\begin{figure*}[t]
	\centering
	\includegraphics[width=0.98\linewidth]{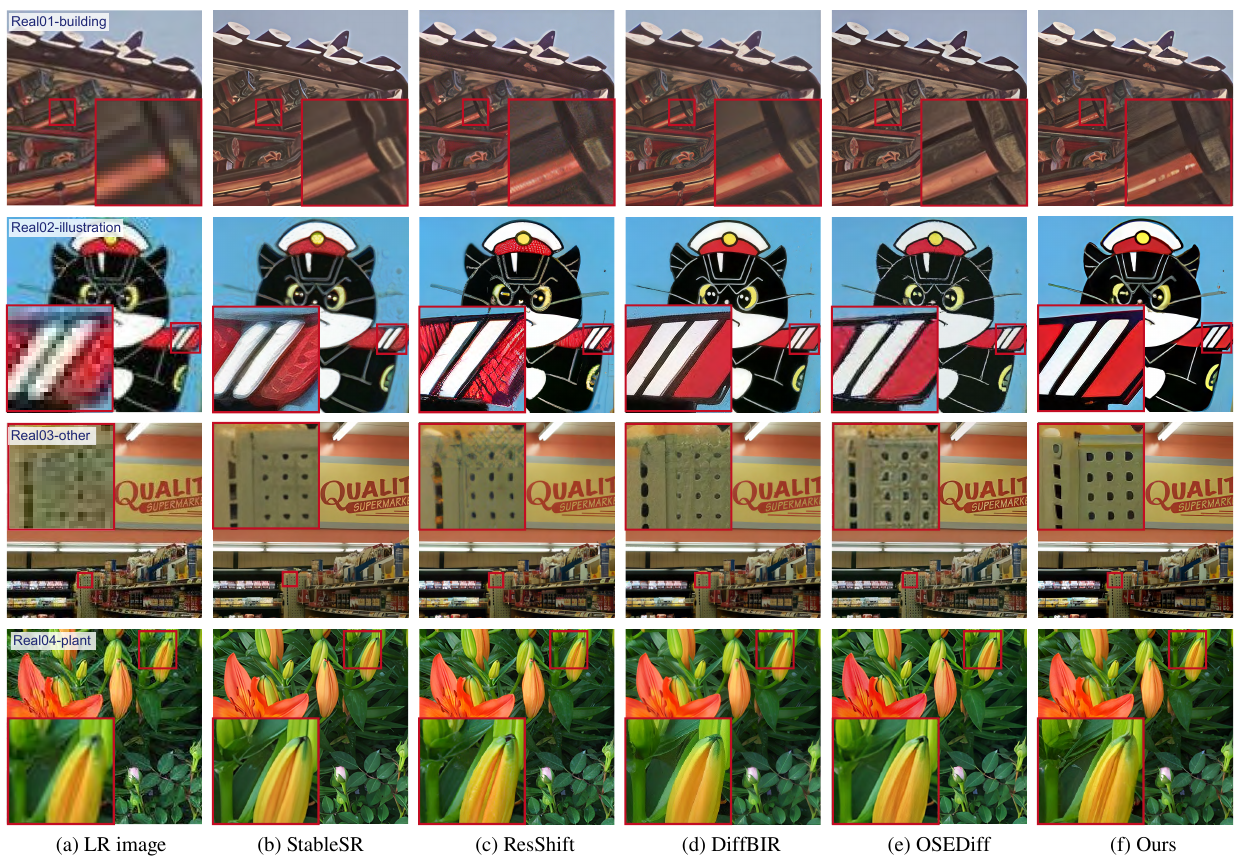}
 \vspace{-2mm}
	\caption{Visual comparisons of state-of-the-art SR methods on real-world low-resolution images.} 
	\label{fig_results_real} 
	\vspace{-4mm}
\end{figure*}
\subsection{Comparisons with State-of-the-Art Methods}
To verify the effectiveness of our PatchScaler, we conduct a series of quantitative and qualitative experiments with the following state-of-the-art SR methods: (1) classic CNN-based and Transformer-based SR methods: RealSR-JPEG~\cite{ji2020real}, BSRGAN~\cite{zhang2021designing}, RealESRGAN~\cite{wang2021real}, and SwinIR~\cite{liang2021swinir}; (2) representative diffusion-based multi-step SR methods: LDM~\cite{rombach2022high}, StableSR~\cite{wang2024exploiting}, ResShift~\cite{yue2024resshift}, DiffBIR~\cite{lin2023diffbir}, and PASD~\cite{yang2023pixel}; (3) the latest diffusion-based one-step SR methods: SinSR~\cite{wang2023sinsr} and OSEDiff~\cite{wu2024one}. 
Following the default settings, the sampling step of ResShift~\cite{yue2024resshift} is set to 15. For other multi-step diffusion-based SR methods, it is set to 20. 
Moreover, we further analyze the performance of SUPIR~\cite{yu2024scaling} and PatchScaler in the supplementary material. 
\begin{table*}[t]
	\centering
	\caption{Running time comparisons of the proposed PatchScaler to other methods on the $\times$ 4 (512 $ \to  $2048) SR task. The results are evaluated using an NVIDIA Tesla A100 GPU. Note that SinSR~\cite{wang2023sinsr} suffers from an out-of-memory problem at this scale. The \textcolor{red!60!black}{\textbf{highlighted}} result is evaluated using dual-GPU parallel computing.}
	\label{tab_runtime_comparison}
	\small
	\vspace{-2mm}
	\resizebox{\textwidth}{!}{
		\begin{tabular}{c|cccccccc|c}
			\Xhline{0.8pt}
			Methods &SwinIR-GAN~\cite{liang2021swinir}  & LDM~\cite{rombach2022high}  & StableSR~\cite{wang2024exploiting}  & ResShift~\cite{yue2024resshift} & DiffBIR~\cite{lin2023diffbir} & PASD~\cite{yang2023pixel} & SinSR~\cite{wang2023sinsr} & OSEDiff~\cite{wu2024one} & PatchScaler \\
			\Xhline{0.4pt}
			Runtime (s) & 1.55  & 25.82  & 35.22  & 14.89  & 78.12  & 59.03 & - & 30.35 & 3.41 (\textcolor{red!60!black}{\textbf{2.26}})       \\
			\Xhline{0.8pt}
	\end{tabular} }
	\vspace{-4mm}
\end{table*}

\begin{figure}[t]
	\centering
	\resizebox{0.27\textwidth}{!}{
		\includegraphics[width=0.30\textwidth]{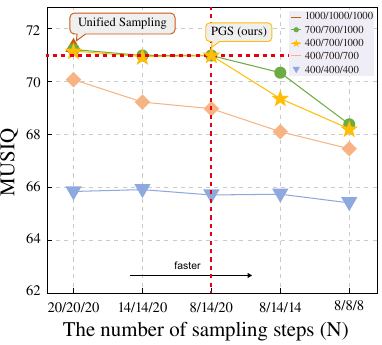}
	}
	\hfill 
		\raisebox{54pt}{
			\begin{minipage}[c]{0.18\textwidth}
				\caption{Performance analysis of PatchScaler on RealSet110 dataset under different configurations. Our PatchScaler achieves a better tradeoff between metrics and runtime at the intersection of two \textcolor{red}{red} dashed lines.}
				\label{fig_configurations}
			\end{minipage}
		}
		\vspace{-5mm}
\end{figure}

\noindent\textbf{Evaluations on Synthetic Data.} 
We first evaluate our PatchScaler on the synthetic LSDIR-TEST dataset. The quantitative results are presented in \Cref{tab_quantitative_results}. 
It can be seen that our method outperforms all others in the non-reference metrics, i.e., ManIQA~\cite{yang2022maniqa}, CLIPIQA~\cite{wang2023exploring} and MUSIQ~\cite{ke2021musiq}. 
ManIQA \cite{yang2022maniqa} utilizes a multi-dimensional attention network for perceptual assessment, 
CLIPIQA \cite{wang2023exploring} leverages rich visual language priors from CLIP to evaluate both the perceptual quality and perceptual abstraction, and MUSIQ \cite{ke2021musiq} captures image quality at different granularities using a multi-scale image quality Transformer. 
The consistent improvements across these metrics highlight PatchScaler’s superior ability to restore fine texture detail and achieve high perceptual quality.  
In addition, the full-reference metrics (i.e., PSNR, SSIM, and LPIPS \cite{zhang2018unreasonable}) are also given in \Cref{tab_quantitative_results} as a reference. 
However, it should be noted that these metrics only reflect certain aspects of performance and are poorly respond to realistic visuals\cite{ledig2017photo, yu2024scaling, blau2018perception, jinjin2020pipal, gu2022ntire}, which cannot reliably evaluate the performance of diffusion-based SR models. 
As the generative capabilities of models improve, it becomes increasingly necessary to reexamine these full-reference metrics and consider more effective evaluation methods for diffusion-based SR models. 
\noindent\textbf{Evaluations on Real-World Data.} 
As shown in \Cref{tab_quantitative_results}, we further evaluate PatchScaler on two real-world datasets. 
It can be seen that our method surpasses others in both ManIQA \cite{yang2022maniqa} and MUSIQ \cite{ke2021musiq} metrics, indicating the excellent performance of our approach. 
Moreover, PatchScaler also demonstrates competitive performance in CLIPIQA \cite{wang2023exploring} metric, with an improvement of \textbf{+0.0204} over the second-best results on the RealSet110 dataset. 
We also provide qualitative comparisons with other diffusion-based SR methods on four LR images in \Cref{fig_results_real}. 
It can be observed that other diffusion-based SR methods struggle to reconstruct realistic textures. 
StableSR~\cite{wang2024exploiting} tends to produce highly blurred (e.g., Real02-illustration) results with unclear texture details. 
The results of ResShift~\cite{yue2024resshift} and DiffBIR~\cite{lin2023diffbir} contain inaccurate generation, which negatively impacts visual perception. 
In addition, OSEDiff~\cite{wu2024one} employs a uniform one-step strategy to minimize sampling steps indiscriminately across all image regions, resulting in unsatisfactory reconstruction results in regions with complex textures (e.g., Real03-other). 
In contrast, our PatchScaler succeeds in producing realistic results at both the patch and global levels, with much better perceptual quality and richer fine-details. 
More visual comparisons can be found in the supplementary material. 
\noindent\textbf{Efficiency Evaluations.} 
To assess the efficiency of PatchScaler, we compare its runtime against other diffusion-based SR methods on the $\times$ 4 (512 $ \to  $2048) SR task, as shown in \Cref{tab_runtime_comparison}. 
We also provide the runtime of the traditional transformer-based method SwinIR-GAN~\cite{liang2021swinir} as a reference. 
Due to the lack of optimization for high-resolution inputs, the inference time of OSEDiff~\cite{wu2024one} is quite long. 
Moreover, SinSR~\cite{wang2023sinsr} suffers from an out-of-memory problem at this scale. 
In contrast, PatchScaler achieves the fastest inference time among the diffusion-based methods, just \textbf{0.23$\times$} that of the second fastest ResShift \cite{yue2024resshift}. 
The efficiency gains of PatchScaler can be attributed to two main factors. 
First, the denoising process in PatchScaler is performed at the patch level, reducing the computational load on the self-attention mechanism across the entire image. 
Second, the PGS treats different image regions discriminatively and dynamically assigns sampling configurations for different patches, providing an efficient inference process, especially for high-resolution images. 
In addition, we highlight that PGS also allows the use of multiple GPUs to process patches from different groups separately. 
To demonstrate this, we re-evaluate our PatchScaler using two NVIDIA Tesla A100 GPUs. 
By assigning one GPU to handle patches from the hard group and another to handle patches from the simple and medium groups, our PatchScaler achieves an additional \textbf{33.7\%} inference speedup over the single-GPU setting, showcasing the scalability of our PatchScaler in parallel computing. 
\subsection{Evaluations of Patch-adaptive Group Sampling}
\begin{figure}[!t]
	\centering
	\includegraphics[width=0.45\textwidth]{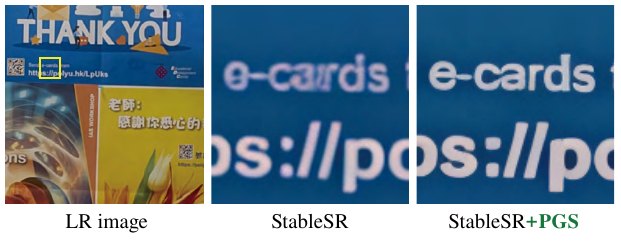}
 \vspace{-2mm}
	\caption{Visual comparison between StableSR and StableSR+PGS. Our method can be easily extended to other baselines and can significantly improve the reconstruction quality.} 
	\label{fig_stablesr} 
	\vspace{-4mm}
\end{figure}

A key feature of PatchScaler is the proposed PGS, which accelerates inference by introducing adaptive sampling for different patches. 
Here, we conduct a series of experiments to analyze the performance of PatchScaler under different configurations and to further validate the extension of PGS. 
\noindent\textbf{Determine Appropriate Configurations.} 
\Cref{fig_configurations} provides an intuitive experiment analysis (metrics against runtime) of the performance of PatchScaler under different configurations. 
Specifically, it can be found that when reducing the intermediate time steps $ T_{1} $, $ T_{2} $, and $ T_{3} $ of simple, medium, and hard patches from [1000, 1000, 1000] to [400, 700, 1000], PatchScaler consistently achieves competitive performance. 
This is because simple patches have higher confidence and a smaller distance $ \triangle \boldsymbol{x}_{0} $ to the ground truth, allowing a larger $ \sqrt{\bar{\alpha}_{\tau}} $ (i.e., a smaller $ \tau $) to satisfy Equation \eqref{pas_diffusion3}, and vice versa. 
However, further reducing these time steps significantly deteriorates performance, indicating that small intermediate time steps fail to satisfy Equation \eqref{pas_diffusion3}. 
Thus, we set $ \tau_{1} $=400, $ \tau_{2} $=700, and $ \tau_{3} $=1000 to achieve an adaptive shortcut path for each group. 
With this configuration, we continue to explore different settings for the sampling step $ N $. 
It can be observed that by reducing the number of iterations $ N_{1} $, $ N_{2} $ and $ N_{3} $ from [20, 20, 20] to [8, 14, 20], the performance of the model can be well maintained. 
This is understandable since a large number of sampling steps is redundant for a small $\tau$. 
However, further reducing the number of sampling steps significantly decreases the performance scores of the three metrics. 
Consequently, there is a trade-off between the sampling step $ N $ and the performance on the setting of $ N_{1} $=8, $ N_{2} $=14, and $ N_{3} $=20. 
By discriminatively assigning sampling steps to each group, PatchScaler achieves an adaptive sampling process with fewer total steps than unified sampling, resulting in a significant speedup the inference process. 
In addition, to better distinguish patches of varying difficulty, we set $\gamma_1$ and $\gamma_2$ to 0.95 and 0.75, respectively, in practice. 
More ablation studies can be found in the supplementary material. 
We present the examples of coarse HR images and corresponding quantified confidence maps $Qmap$ in \Cref{fig_pas}. 
It can be seen that this setting can accurately quantify the reconstruction difficulty of different regions. 
\noindent\textbf{The Extension of PGS.} 
Since PGS is not involved in model training, it can be seamlessly extended to other architectures and tasks as a plug-and-play component and achieve dynamic acceleration of model inference. 
As an example, we applied the proposed PGS and GRM to StableSR~\cite{wang2024exploiting} in a trainable-free manner. 
Since its backbone, Stable Diffusion~\cite{rombach2022high}, performs best with a latent feature size of 64, we set the patch size \textit{V} to 64. 
The quantitative and qualitative results are presented in \Cref{tab_study_stablesr} and \Cref{fig_stablesr}, which show that our PGS achieves significant improvements in both quantitative metrics (ManIQA: \textbf{+0.1179}; MUSIQ:\textbf{+3.46}) and qualitative comparisons, and further accelerates the model inference efficiency (only \textbf{0.24$\times$}) of StableSR. 
This reveals that adopting a unified sampling strategy is suboptimal and redundant. 
In contrast, by dynamically assigning sampling configurations to different patches, our PGS can significantly accelerate the inference process of existing diffusion-based models, While maintaining reconstruction quality. 
\begin{table}[t]
	\centering
	\caption{Quantitative comparison of StableSR and StableSR+PGS. The best results are \textcolor{red!60!black}{\textbf{highlighted}}. The runtime is measured on the $ \times $4 (512 $ \to  $2048) SR task using an NVIDIA Tesla A100 GPU.}
	\label{tab_study_stablesr}
	\small
	\vspace{-2mm}
        \resizebox{0.48\textwidth}{!}{
	\begin{tabular}{c|ccc|c}
		\Xhline{0.8pt}
		Methods & ManIQA & CLIPIQA & MUSIQ & Runtime (s)      \\ 
		\hline
		StableSR & 0.3856 & \textcolor{red!60!black}{\textbf{0.7851}}  & 62.08 & 35.22 \\
        StableSR+PGS & \textcolor{red!60!black}{\textbf{0.5035}} & 0.7809  & \textcolor{red!60!black}{\textbf{65.54}} & \textcolor{red!60!black}{\textbf{8.53}} \\
		\Xhline{0.8pt}
	\end{tabular} }
\end{table}
\begin{table}[t]
	\centering
	\caption{Ablation studies of different prompt in our method. The best results are \textcolor{red!60!black}{\textbf{highlighted}}.}
	\label{tab_study_prompt}
	\small
	\vspace{-2mm}
	\begin{tabular}{@{}C{2.5cm}@{}|@{}C{1.7cm}@{} @{}C{1.7cm}@{} @{}C{1.7cm}@{}}
		\Xhline{0.8pt}
		Prompt type & ManIQA & CLIPIQA & MUSIQ      \\ 
		\hline
		Text prompt & 0.4849 & 0.7890  & 66.79  \\
        Texture prompt & \textcolor{red!60!black}{\textbf{0.5442}} & \textcolor{red!60!black}{\textbf{0.8213}}  & \textcolor{red!60!black}{\textbf{70.97}}  \\
		\Xhline{0.8pt}
	\end{tabular} 
 \vspace{-3mm}
\end{table}
\subsection{Text Prompt vs. Texture Prompt}
To further improve the denoising capability at each step of model inference, we introduce a texture prompt for Patch-DiT, which mitigates the text-image misalignment problem associated with text prompt and provides rich conditional information by retrieving high-quality and patch-independent texture priors from a reference texture memory. 
In this section, we conduct ablation studies between text prompt and texture prompt on the RealSet110 dataset. 
Since HR images are not available during inference, we utilize the BLIP-2 vision-language pre-training model \cite{li2023blip} to generate the text information from the LR image, and then obtain the text prompt based on the CLIP model \cite{radford2021learning}. 
The results are presented in \Cref{tab_study_prompt}. 
It can be seen that the model with a texture prompt achieves higher scores in three quantitative metrics than the model with a text prompt. 
This is due to the challenging misalignment~\cite{chen2023pixart} between the text content and image content in the SISR task, which often leads to degraded performance. 
We discuss this misalignment problem further in the supplementary material by analyzing the performance of SUPIR~\cite{yu2024scaling} and PatchScaler. 
These conclusions demonstrate the effectiveness of the proposed texture prompt in improving the performance of PatchScaler. 
\subsection{Discussion}
Since the unified sampling process with a large number of steps is applied indiscriminately to different image regions, the existing diffusion-based multi-step SR methods (e.g., DiffBIR~\cite{lin2023diffbir}, PASD~\cite{yang2023pixel}, and SUPIR~\cite{yu2024scaling}) require expensive computational costs. 
This restricts their applicability to a few non-time-sensitive post-processing scenarios. 
Although real-time performance remains unresolved for all diffusion-based SR methods, our PatchScaler shows a significant improvement in computational efficiency while maintaining reconstruction quality, particularly when dealing with high-resolution images. 
In the quantitative evaluation of \Cref{tab_runtime_comparison}, the runtime of PatchScaler outperforms diffusion-based one-step SR methods and rivals state-of-the-art GAN-based models (e.g. SwinIR-GAN~\cite{liang2021swinir}). 
This presents exciting opportunities for a wide range of users, including researchers, photographers, and technology companies. 
In the future, we will explore integration options with existing sampling distillation solutions and further improve the deployment efficiency through engineering solutions. 

%

%
\section{Conclusion}
 In this paper, we have presented PatchScaler, an efficient patch-independent diffusion-based SR model. It is motivated by the observation that performing the unified sampling  equally on different images regions is redundant. 
 We developed a PGS strategy to discriminatively assign an appropriate sampling configuration for each patch according to a quantified confidence map so that the high-resolution images can be efficiently recovered. 
 Additionally, we introduced a texture prompt for Patch-DiT to provide rich texture condition information by retrieving high-quality patch-independent texture priors from a reference texture memory. 
 Our PatchScaler can efficiently solve SISR problem and achieve superior performance both quantitatively and qualitatively. 

{
	\small
	\bibliographystyle{ieeenat_fullname}
	\bibliography{main}
}

\end{document}